Elias Bareinboim* and Judea Pearl

# A General Algorithm for Deciding Transportability of Experimental Results

**Abstract:** Generalizing empirical findings to new environments, settings, or populations is essential in most scientific explorations. This article treats a particular problem of generalizability, called "transportability", defined as a license to transfer information learned in experimental studies to a different population, on which only observational studies can be conducted. Given a set of assumptions concerning commonalities and differences between the two populations, Pearl and Bareinboim [1] derived sufficient conditions that permit such transfer to take place. This article summarizes their findings and supplements them with an effective procedure for deciding when and how transportability is feasible. It establishes a necessary and sufficient condition for deciding when causal effects in the target population are estimable from both the statistical information available and the causal information transferred from the experiments. The article further provides a complete algorithm for computing the transport formula, that is, a way of combining observational and experimental information to synthesize bias-free estimate of the desired causal relation. Finally, the article examines the differences between transportability and other variants of generalizability.

*Corresponding author: **Elias Bareinboim,** Department of Computer Science, University of California, Los Angeles, CA, USA, E-mail: eb@cs.ucla.edu
**Judea Pearl,** Department of Computer Science, University of California, Los Angeles, CA, USA, E-mail: judea@cs.ucla.edu

# 1 Introduction

The problem of transporting knowledge from one population to another is pervasive in science. Conclusions that are obtained in a laboratory setting are transported and applied elsewhere, in an environment that differs in many aspects from that of the laboratory. Experiments conducted on a group of subjects are intended to inform policies on a different group, usually more general and in which the studied group is just one of its parts.

Surprisingly, the conditions under which this extrapolation can be legitimized were not formally articulated until very recently [1–3]. Although the problem has been discussed in many areas of statistics, economics, and the health sciences, under rubrics such as "external validity" [4, 5], "meta-analysis" [6–8], "overgeneralization" [9], "quasi experiments" [10, 11 (Ch. 3)], "heterogeneity" [12], these discussions are limited to verbal narratives in the form of heuristic guidelines for experimental researchers – no formal treatment of the problem has been attempted to answer the practical problem of generalizing across populations posed in this article. (See Section 6 for related work.)

Recent developments in causal inference enable us to tackle this problem formally. First, the distinction between statistical and causal knowledge has received syntactic representation through causal diagrams [13–16]. Second, graphical models provide a language for representing differences and commonalities among domains, environments, and populations [1]. Finally, the inferential machinery provided by the do-calculus [13, 16, 17] is particularly suitable for combining these two advances into a coherent framework and developing effective algorithms for knowledge transfer.

Armed with these tools, we consider transferring causal knowledge between two populations $\Pi$ and $\Pi^*$. In population $\Pi$, experiments can be performed and causal knowledge gathered. In $\Pi^*$, potentially different from $\Pi$, only passive observations can be collected but no experiments conducted. The problem is to infer a causal relationship $R$ in $\Pi^*$ using knowledge obtained in $\Pi$. Clearly, if nothing is known about the relationship between $\Pi$ and $\Pi^*$, the problem is trivial; no transfer can be justified. Yet the fact that all experiments are conducted with the intent of being used elsewhere (e.g., outside the laboratory) implies that scientific explorations are driven by the assumption that certain populations share common characteristics and that, owed to these commonalities, causal claims would be valid in new settings even where experiments cannot be conducted.

To formally articulate commonalities and differences between populations, a graphical representation named *selection diagrams* was devised in [1], which represent differences in the form of unobserved factors capable of causing such differences. Given an arbitrary selection diagram, our challenge is to decide whether commonalities override differences to permit the transfer of information across the two populations. We show that this challenge can be met by an effective procedure that decides when and how transportability is feasible.

The article is organized as follows. In section 2, we motivate the problem of transportability using three simple examples and informally summarize the findings of Pearl and Bareinboim [1]. In section 3, we formally define the notion of selection diagrams and transportability, exemplify how it can be reduced to a problem of symbolic transformation in do-calculus, and provide examples for models that prohibit transportability. In section 4, we provide a graphical criterion for deciding transportability in arbitrary diagrams. In section 5, we provide an effective procedure for deciding transportability, which returns a correct transport formula whenever such exists. In section 6, we compare transportability to other problems of generalizing empirical findings. Section 7 provides concluding remarks.

## 2 Motivation

To motivate the formal treatment of transportability, we use three simple examples taken from [1] and graphically depicted in Figure 1.

**Example 1.** *Consider the problem of transferring experimental results between two locations. We first conduct a randomized trial in Los Angeles (LA) and estimate the causal effect of treatment $X$ on outcome $Y$ for every age group $Z = z$, denoted $P(y|do(x),z)$. We now wish to generalize the results to the population of New York City (NYC), but we find the distribution $P(x,y,z)$ in LA to be different from the one in NYC (call the latter $P^*(x,y,z)$). In particular, the average age in NYC is significantly higher than that in LA. How are we to estimate the causal effect of $X$ on $Y$ in NYC, denoted $R = P^*(y|do(x))$?*[1]

The selection diagram for this example (Figure 1(a)) conveys the assumption that the *only* difference between the two population are factors determining age distributions, shown as $S \rightarrow Z$, while age-specific effects $P(y|do(x), Z = z)$ are invariant across cities. Difference-generating factors are represented by a special set of variables called *selection variables* $S$ (or simply $S$-variables), which are graphically depicted as square nodes (■).[2] From this assumption, the overall causal effect in NYC can be derived as follows[3]:

$$R = \sum_z P^*(y|do(x),z)P^*(z)$$
$$= \sum_z P(y|do(x),z)P^*(z) \qquad [1]$$

---

[1] We will later on use $P_x(y)$ interchangeably with $P(y|do(x))$.
[2] See Def. 3 below for formal construction of selection diagrams. In all diagrams, dashed arcs (e.g., $X \leftarrow\rightarrow Y$) represent the presence of latent variables affecting both $X$ and $Y$.
[3] This result can be derived by purely graphical operations if we write $P^*(y|do(x),z)$ as $P(y|do(x),z,s)$, thus attributing the difference between $\Pi$ and $\Pi^*$ to a fictitious event $S = s$. The invariance of the age-specific effect then follows from the conditional independence $(S \perp\!\!\!\perp Y | Z, X)_{G_{\bar{x}}}$, which implies $P(y|do(x),z,s) = P(y|do(x),z)$, and licenses the derivation of the transport formula.

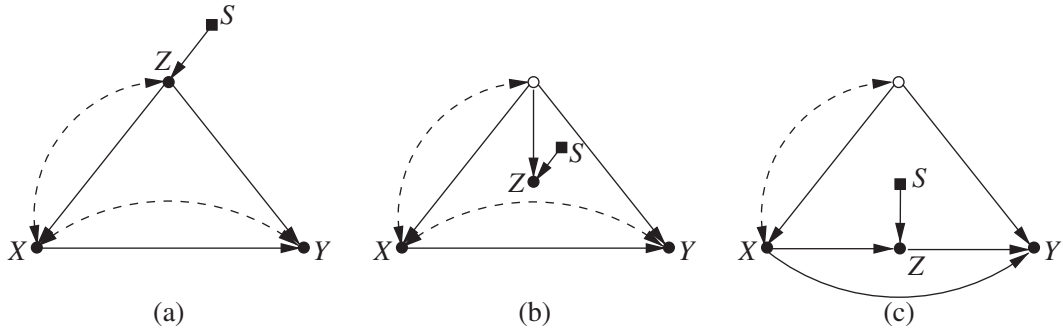

**Figure 1** Causal diagrams depicting Examples 1–3. In (a) Z represents "age." In (b) Z represents "linguistic skills" while age (in hollow circle) is unmeasured. In (c) Z represents a biological marker situated between the treatment (X) and a disease (Y).

The last line constitutes a *transport formula* for R. It combines experimental results obtained in LA, $P(y|do(x),z)$, with observational aspects of NYC population, $P^*(z)$, to obtain an experimental claim $P^*(y|do(x))$ about NYC.[4]

Our first task in this article will be to explicate the assumptions that renders this extrapolation valid. We ask, for example, what must we assume about other confounding variables beside age, both latent and observed, for eq. [1] to be valid, or, would the same transport formula hold if Z was not age, but some proxy for age, say, "language skills" (Figure 1(b)). More intricate yet, what if Z stood for an exposure-dependent variable, say hyper-tension level, that stands between X and Y (Figure 1(c))?

Let us examine the proxy issue first.

**Example 2.** *Let the variable Z in Example 1 stand for subjects' language skills, and let us assume that Z does not affect exposure (X) or outcome (Y), yet it correlates with both, being a proxy for age which is not measured in either study (see Figure 1(b)). Given the observed disparity $P(z) \neq P^*(z)$, how are we to estimate the causal effect $P^*(y|do(x))$ for the target population of NYC from the z-specific causal effect $P(y|do(x),z)$ estimated at the study population of LA?*

Our intuition dictates, and correctly so, that since reading ability has no causal effect on treatment nor on the outcome the proper transport formula would be

$$P^*(y|do(x)) = P(y|do(x)) \quad [2]$$

namely, the causal effect is "directly" transportable with no calibration needed (to be shown later on). This will be the case even if the observed joint distribution $P^*(x,y,z)$ is the same as in Example 1 where Z stands for age. We see, therefore, that the proper transport formula depends on the causal context in which population differences are embedded, not merely on the joint distribution over the observed variables.

This example also demonstrates why the invariance of Z-specific causal effects should not be taken for granted. While justified in Example 1, with Z = age, it fails in Example 2, in which Z was equated with "language skills." The intuition is clear. A NYC person at skill level $Z = z$ is likely to be in a totally different age group from his skill-equals in LA and, since it is age, not skill that shapes the way individuals respond to treatment, it is only reasonable that LA residents would respond differently to treatment than their NYC counterparts at the very same skill level.

**Example 3.** *Examine the case where Z is a X-dependent variable, say a disease bio-marker, standing on the causal pathways between X and Y as shown in Figure 1(c). Assume further that the disparity $P(z) \neq P^*(z)$ is*

---

[4] Eq. [1] reflects the familiar method of "standardization" – a statistical extrapolation method that can be traced back to a century-old tradition in demography and political arithmetic [18–21]. We will show that standardization is only valid under certain conditions.

*discovered in each level of X and that, again, both the average and the z-specific causal effect $P(y|do(x), z)$ are estimated in the LA experiment, for all levels of X and Z. Can we, based on the information given, estimate the average (or z-specific) causal effect in the target population of NYC?*

Assuming that the disparity in $P(z)$ stems only from a difference in subjects' susceptibility to $X$, as encoded in the selection the diagram of Figure 1(c), we will demonstrate in section 3 that the correct transport formula should be

$$P^*(y|do(x)) = \sum_z P(y|do(x), z) P^*(z|x), \quad [3]$$

which is different from both eqs. [1] and [2]. It calls instead for the z-specific effects to be weighted by the conditional probability $P^*(z|x)$, estimated at the target population.

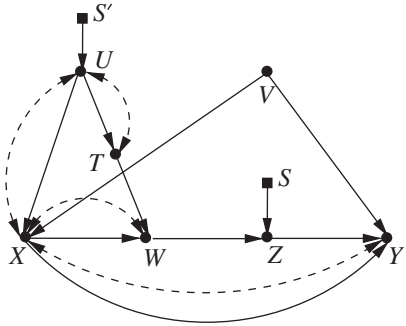

**Figure 2** Selection diagram with two "difference-producing" factors ($S$ and $S'$); the derivation of transportability is more involved using Lemma 1, and it is shown step by step using the algorithm in section 5.

In these three intuitive examples transportability amounts to simple operations (i.e., recalibration, direct transport, and weighted recalibration); however, in more elaborate examples, the full power of formal analysis would be required. For instance, Pearl and Bareinboim [1] showed that, in the problem depicted in Figure 2, where both the Z-determining mechanism and the U-determining mechanism are suspect of being different, the transport formula for the relation $P^*(y|do(x))$ is given by

$$\sum_z P(y|do(x), z) \sum_w P^*(z|w) \sum_t P(w|do(x), t) P^*(t)$$

This formula instructs us to estimate $P(y|do(x), z)$ and $P(w|do(x), t)$ in the experimental population, then combine them with the estimates of $P^*(z|w)$ and $P^*(t)$ in the target population. Pearl and Bareinboim [1] derived this formula using the following lemma, which translates the property of transportability to the existence of a syntactic reduction using a sequence of do-calculus rules.

**Lemma 1** [1]. *Let D be the selection diagram characterizing $\Pi$ and $\Pi^*$, and S a set of selection variables in D. The relation $R = P^*(y|do(x), z)$ is transportable from $\Pi$ to $\Pi^*$ if the expression $P(y|do(x), z, s)$ is reducible, using the rules of do-calculus, to an expression in which S appears only as a conditioning variable in do-free terms.*

The logic of this reduction is simple. Terms lacking an S variable are estimable at the source population while those lacking the do-operator are estimable non-experimentally at the target population. If such a reduction exists, the resulting expression gives the transport formula for R.

Lemma 1 is declarative but not computationally effective, for it does not specify the sequence of rules leading to the needed reduction, nor does it tell us if such a sequence exists. It is useful primarily as a verification tool, to confirm the transportability of a given relation once we are in possession of a "witness" sequence.

To overcome this deficiency, Pearl and Bareinboim [1] proposed a recursive procedure (their Theorem 3), which can handle many cases, among them Figure 2, but is not "complete", that is, diagrams exist that support transportability and which the recursive procedure fails to recognize as such. The procedure developed in this article are guaranteed to make correct identification in all cases. We summarize our contributions as follows:

- We derive a general graphical condition for deciding transportability of causal effects. We show that transportability is feasible if and only if a certain graph structure does not appear as an edge subgraph of the inputted selection diagram.
- We provide necessary or sufficient graphical conditions for special cases of transportability, for instance, controlled direct effects (CDE).
- We construct a complete algorithm for deciding transportability of joint causal effects and returning a proper transport formula whenever those effects are transportable.

## 3 Preliminaries

The semantical framework in our analysis rests on *structural causal models* (SCM) as defined next, also called *probabilistic causal models* or *data-generating models*.

**Definition 1** (Structural Causal Model [22, p. 203]). *A SCM is a 4-tuple $M = \langle U, V, F, P \rangle$ where:*

1. *$U$ is a set of background or exogenous variables, representing factors outside the model, which nevertheless affect relationships within the model.*
2. *$V$ is a set of endogenous variables $\{V_1, ..., V_n\}$, assumed to be observable. Each of these variables is functionally dependent on some subset $PA_i$ of $U \cup V \setminus \{V_i\}$.*
3. *$F$ is a set of functions $\{f_1, ..., f_n\}$ such that each $f_i$ determines the value of $V_i \in V$, $v_i = f_i(pa_i, u)$.*
4. *A joint probability distribution $P(u)$ over $U$.*

In the structural causal framework [22, Ch. 7], actions are modifications of functional relationships, and each action $do(x)$ on a causal model $M$ produces a new model $M_x = \langle U, V, F_x, P(U) \rangle$, where $F_x$ is obtained after replacing $f_X \in F$ for every $X \in X$ with a new function that outputs a constant value $x$ given by $do(x)$. See Appendix 1 for a gentle introduction to structural models, or [23] for a more detailed discussion.

We follow the conventions given in [22]. We will denote variables by capital letters and their values by small letters. Similarly, sets of variables will be denoted by bold capital letters, sets of values by bold letters. We will use the typical graph-theoretic terminology with the corresponding abbreviations $Pa(Y)_G$, $An(Y)_G$, and $De(Y)_G$, which will denote respectively the set of observable parents, ancestors, and descendants of the node set $Y$ in $G$. By convention, these sets will include the arguments as well, for instance, the ancestral set $An(Y)_G$ will include $Y$. We will usually omit the graph subscript whenever the graph in question is assumed or obvious. A graph $G_Y$ will denote the induced subgraph $G$ containing nodes in $Y$ and all arrows between such nodes. Finally, $G_{\overline{X}\underline{Z}}$ stands for the edge subgraph of $G$ where all incoming arrows into $X$ and all outgoing arrows from $Z$ are removed.

Key to the analysis of transportability is the notion of "identifiability," defined below, which expresses the requirement that causal effects be computable from a combination of data $P$ and assumptions embodied in a causal graph $G$.

**Definition 2** (Causal Effects Identifiability [22, p. 77]). *The causal effect of an action $do(x)$ on a set of variables $Y$ such that $Y \cap X = \emptyset$ is said to be identifiable from $P$ in $G$ if $P_x(y)$ is uniquely computable from $P(V)$ in any model that induces $G$.*

Causal models and their induced graphs are normally associated with one particular domain (also called setting, study, population, environment). In the transportability case, we extend this representation to capture properties of several domains simultaneously. This is made possible if we assume that there are no structural changes between the domains, that is, all structural equations share the same set of arguments, though the functional forms of the equations may vary arbitrarily.[5,6]

**Definition 3** (Selection Diagram). *Let $\langle M, M^* \rangle$ be a pair of SCM relative to domains $\langle \Pi, \Pi^* \rangle$, sharing a causal diagram G. $\langle M, M^* \rangle$ is said to induce a selection diagram D if D is constructed as follows:*

1. *Every edge in G is also an edge in D;*
2. *D contains an extra edge $S_i \rightarrow V_i$ whenever there might exist a discrepancy $f_i \neq f_i^*$ or $P(U_i) \neq P^*(U_i)$ between M and $M^*$.*

In words, the *S*-variables locate the *mechanisms* where structural discrepancies between the two domains are suspected to take place.[7] Alternatively, one can see a selection diagram as a carrier of invariance claims between the mechanisms of both domains – the absence of a selection node pointing to a variable represents the assumption that the mechanism responsible for assigning value to that variable is the same in the two domains.[8]

Armed with a selection diagram and the concept of identifiability, transportability of causal effects (or transportability, for short) can be defined as follows:

**Definition 4** (Causal Effects Transportability). *Let D be a selection diagram relative to domains $\langle \Pi, \Pi^* \rangle$. Let $\langle P, I \rangle$ be the pair of observational and interventional distributions of $\Pi$, and $P^*$ be the observational distribution of $\Pi^*$. The causal effect $R = P_{\mathbf{x}}^*(\mathbf{y})$ is said to be transportable from $\Pi$ to $\Pi^*$ in D if $P_{\mathbf{x}}^*(\mathbf{y})$ is uniquely computable from $P, P^*, I$ in any model that induces D.*

In some broad sense, one can view transportability as a special case of identifiability, where the pair of structures constitutes a global model, and the task is to infer a property of one population from sum total of the information available (i.e., $\langle P, I, P^* \rangle$). However, the unique challenges of dealing with two diverse environments under two different experimental regimes, and the special problems that emerge from this combination can benefit appreciably from viewing transportability as distinct major extension of identifiability. To witness, all identifiable causal relations in $(G^*, P^*)$ are also transportable, because they can be computed directly from $\Pi^*$ and require no experimental information from $\Pi$. This observation engender the following definition of *trivial transportability*.

**Definition 5** (Trivial Transportability). *A causal relation R is said to be trivially transportable from $\Pi$ to $\Pi^*$, if $R(\Pi^*)$ is identifiable from $(G^*, P^*)$.*

The following observation establishes another connection between identifiability and transportability. For a given causal diagram G, one can produce a selection diagram D such that identifiability in G is equivalent to transportability in D. First set $D = G$, and then add selection nodes pointing to all variables in D, which

---

**5** This definition was left implicit in [1].
**6** The assumption that there are no structural changes between domains can be relaxed as follows. Starting with the structure in the target population $G^*$, make $D = G^*$, and then add S-nodes to D following the same procedure as in Def. 3.
**7** Transportability analysis assumes that enough structural knowledge about both domains is known in order to substantiate the production of their respective causal diagrams. In the absence of such knowledge, *causal discovery* algorithms might be used to help in inferring the diagrams from data [15, 22, 24].
**8** These invariance assumptions are analogous to the missing-arrows in the causal graphs [25] which allow one to identify causal-effects from observational data.

represents that the target domain does not share any commonality with its pair – this is equivalent to the problem of identifiability because the only way to achieve transportability is to identify $R$ from scratch in the target domain.

Another special case of transportability occurs when a causal relation has identical form in both domains – no recalibration is needed. This is captured by the following definition.

**Definition 6** (Direct Transportability). *A causal relation $R$ is said to be directly transportable from $\Pi$ to $\Pi^*$, if $R(\Pi^*) = R(\Pi)$.*

A graphical test for direct transportability of $R = P^*(y|do(x), z)$ follows from do-calculus and reads: $(S \perp\!\!\!\perp Y | X, Z)_{G_{\overline{X}}}$; in words, $X$ blocks all paths from $S$ to $Y$ once we remove all arrows pointing to $X$ and condition on $Z$. As a concrete example, the $z$-specific effect in Figure 1(a) is the same in both domains; hence, it is directly transportable. Also, the effect $P^*(y|do(x))$ in Figure 1(b) is the same in both domains; hence, it is directly transportable.

These two cases will act as a basis to decompose the problem of transportability into smaller and more manageable subproblems. For instance, let us estimate the effect $R = P^*(y|do(x))$ in the bio-marker example depicted in Figure 1(c).

$$P^*(y|do(x)) = \sum_z P^*(y|do(x), z) P^*(z|do(x)) \qquad [4]$$

$$= \sum_z P^*(y|do(x), z) P^*(z|x) \qquad [5]$$

$$= \sum_z P(y|do(x), z) P^*(z|x), \qquad [6]$$

In eq. [4], the target relation $R$ is conditioned on $Z$. The effect $P^*(z|do(x))$ in eq. [5] is trivially transportable since it is identifiable in $\Pi^*$, and $P^*(y|do(x), z)$ in eq. [6] is directly transportable since $(S \perp\!\!\!\perp Y|X, Z)_{G_{\overline{X}}}$.

Now we turn our attention to conditions that preclude identifiability. The following lemma provides an auxiliary tool to prove non-transportability and is based on refuting the uniqueness property required by Definition 4.

**Lemma 2**. *Let $\mathbf{X}, \mathbf{Y}$ be two sets of disjoint variables, in population $\Pi$ and $\Pi^*$, and let $D$ be the selection diagram. $P^*_\mathbf{x}(\mathbf{y})$ is not transportable from $\Pi$ to $\Pi^*$ if there exist two causal models $M^1$ and $M^2$ compatible with $D$ such that $P_1(\mathbf{V}) = P_2(\mathbf{V})$, $P^*_1(\mathbf{V}) = P^*_2(\mathbf{V})$, $P_1(\mathbf{V}\backslash\mathbf{W}|do(\mathbf{W})) = P_2(\mathbf{V}\backslash\mathbf{W}|do(\mathbf{W}))$, for any set $\mathbf{W}$, all families have positive distribution, and $P^*_1(\mathbf{y}|do(\mathbf{x})) \neq P^*_2(\mathbf{y}|do(\mathbf{x}))$.*

*Proof.* Let $I$ be the set of interventional distributions $P(\mathbf{V}\backslash\mathbf{W}|do(\mathbf{W}))$, for any set $W$. The latter inequality rules out the existence of a function from $P, P^*, I$ to $P^*_\mathbf{x}(\mathbf{y})$. ∎

While the problems of identifiability and transportability are related, Lemma 2 indicates that proofs of non-transportability are more involved than those of non-identifiability. Indeed, to prove non-transportability requires the construction of two models agreeing on $\langle P, I, P^* \rangle$, while non-identifiability requires the two models to agree solely on the observational distribution $P$.

The simplest non-transportable structure is an extension of the famous "bow arc" graph named here "s-bow arc," see Figure 3(a). The s-bow arc has two endogenous nodes: $X$, and its child $Y$, sharing a hidden exogenous parent $U$, and a $S$-node pointing to $Y$. This and similar structures that prevent transportability will be useful in our proof of completeness, which requires a demonstration that whenever the algorithm fails to transport a causal relation, the relation is indeed non-transportable.

**Theorem 1.** *$P^*_x(y)$ is not transportable in the s-bow arc graph.*

*Proof.* The proof will show a counterexample to the transportability of $P^*_x(Y)$ through two models $M_1$ and $M_2$ that agree in $\langle P, P^*, I \rangle$ and disagree in $P^*_x(y)$.

Assume that all variables are binary. Let the model $M_1$ be defined by the following system of structural equations: $X_1 = U, Y_1 = ((X \otimes U) \otimes S), P_1(U) = 1/2$, and $M_2$ by the following one: $X_2 = U, Y_2 = S \vee (X \otimes U), P_2(U) = 1/2$, where $\otimes$ represents the *exclusive or* function.

**Lemma 3.** *The two models agree in the distributions $\langle P, P^*, I \rangle$.*

*Proof.* We show that the following equations must hold for $M_1$ and $M_2$:

$$\begin{cases} P_1(X|S) = P_2(X|S), & S = \{0,1\} \\ P_1(Y|X,S) = P_2(Y|X,S), & S = \{0,1\} \\ P_1(Y|do(X), S=0) = P_2(Y|do(X), S=0) \end{cases}$$

for all values of $X, Y$. The equality between $P_i(X|S)$ is obvious since $(S \perp\!\!\!\perp X)$ and $X$ has the same structural form in both models. Second, let us construct the truth table for $Y$:

| X | S | U | $Y_1$ | $Y_1$ |
|---|---|---|---|---|
| 0 | 0 | 0 | 0 | 0 |
| 0 | 0 | 1 | 1 | 1 |
| 0 | 1 | 0 | 1 | 1 |
| 0 | 1 | 1 | 0 | 1 |
| 1 | 0 | 0 | 1 | 1 |
| 1 | 0 | 1 | 0 | 0 |
| 1 | 1 | 0 | 0 | 1 |
| 1 | 1 | 1 | 1 | 1 |

To show that the equality between $P_i(Y=1|X, S=0), X = \{0,1\}$ holds, we rewrite it as follows:

$$P_i(Y=1|X, S=0) = \frac{P_i(Y=1|X, S=0, U=1) P_i(X|U=1) P_i(U=1)}{P_i(X)} \\ + \frac{P_i(Y=1|X, S=0, U=0) P_i(X|U=0) P_i(U=0)}{P_i(X)} \quad [7]$$

In eq. [7], the expressions for $X = \{0, 1\}$ are functions of the tuples $\{(X=1, S=0, U=1), (X=0, S=0, U=0)\}$, which evaluate to the same value in both models. Similarly, the expressions $P_i(Y=1|X, S=1)$ for $X = \{0, 1\}$ are functions of the tuples $\{(X=1, S=1, U=1), (X=0, S=1, U=0)\}$, which also evaluate to the same value in both models.

We further assert the equality between the interventional distributions in $\Pi$, which can be written using the do-calculus as

$$P_i(Y=1|do(X), S=0) = \sum_U P_i(Y|do(X), S=0, U) P_i(U|do(X), S=0) \\ = P_i(Y=1|X, S=0, U=1) P_i(U=1) \\ + P_i(Y=1|X, S=0, U=0) P_i(U=0), \quad X = \{0,1\} \quad [8]$$

Evaluating this expression points to the tuples $\{(X=1, S=0, U=1), (X=1, S=0, U=0)\}$ and $\{(X=0, S=0, U=1), (X=0, S=0, U=0)\}$, which map to the same value in both models. ∎

**Lemma 4.** *There exist values of $X, Y$ such that $P_1(Y|do(X), S=1) \neq P_2(Y|do(X), S=1)$.*

*Proof.* Fix $X = 1, Y = 1$, and let us rewrite the desired quantity in $\Pi^*$ as

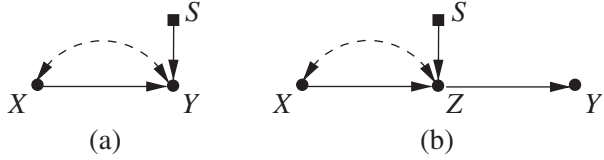

**Figure 3** (a) Smallest selection diagram in which $P^*(y|do(x))$ is not transportable (s-bow graph). (b) A selection diagram in which even though there is no S-node pointing to Y, the effect of X on Y is still not-transportable due to the presence of a sC-tree (see Corollary 2).

$$
\begin{aligned}
P_i(Y=1|do(X=1), S=1) &= \sum_U P_i(Y|do(X=1), S=1, U) P_i(U|do(X=1), S=1) \\
&= P_i(Y=1|X=1, S=1, U=1) P_i(U=1) \\
&\quad + P_i(Y=1|X=1, S=1, U=0) P_i(U=0)
\end{aligned}
\quad [9]
$$

Since $R_i$ is a function of the tuples $\{(X=1, S=1, U=1), (X=1, S=1, U=0)\}$, it evaluates in $M_1$ to $\{1,1\}$ and in $M_2$ to $\{1,0\}$.

Hence, together with the uniformity of $P(U)$, it follows that $R_1 = 1$ and $R_2 = 1/2$, which finishes the proof. ∎

By Lemma 2, Lemmas 3 and 4 prove Theorem 1. ∎

## 4 Characterizing transportable relations

The concept of confounded components (or *C-components*) was introduced in [26] to represent clusters of variables connected through bidirected edges and was instrumental in establishing a number of conditions for ordinary identification (Def. 2). If $G$ is not a *C*-component itself, it can be uniquely partitioned into a set $C(G)$ of *C*-components. We now recast *C*-components in the context of transportability.[9]

**Definition 7** (sC-component). *Let G be a selection diagram such that a subset of its bidirected arcs forms a spanning tree over all vertices in G. Then G is a sC-component (selection confounded component).*

A special subset of *C*-components that embraces the ancestral set of $Y$ was noted by Shpitser and Pearl [27] to play an important role in deciding identifiability – this observation can also be applied to transportability, as formulated in the next definition.

**Definition 8** (sC-tree). *Let G be a selection diagram such that $C(G) = \{G\}$, all observable nodes have at most one child, there is a node Y, which is a descendent of all nodes, and there is a selection node pointing to Y. Then G is called a Y-rooted sC-tree (selection confounded tree).*

The presence of this structure (and generalizations) will prove to be an obstacle to transportability of causal effects. For instance, the s-bow arc in Figure 3(a) is a *Y*-rooted *sC*-tree where we know $P^*_x(y)$ is not transportable there.

---

**9** Departing from results given in [28–32], the advent of C-components complements the notion of *inducing path*, which was earlier introduced in [33], and led to a breakthrough result proving completeness of the *do*-calculus for non-parametric identification of causal effects by [27, 34].

In certain classes of problems, the absence of such structures will prove sufficient for transportability. One such class is explored below and consists of models in which the set $X$ coincides with the parents of $Y$.

**Theorem 2.** *Let G be a selection diagram. Then for any node Y, the causal effects $P^*_{Pa(Y)}(y)$ is transportable if there is no subgraph of G which forms a Y-rooted sC-tree.*

*Proof.* See Appendix 2. ∎

Theorem 2 provides a tractable transportability condition for the CDE – a key concept in modern mediation analysis, which permits the decomposition of effects into their direct and indirect components [35, 36]. CDE is defined as the effect of $X$ on $Y$ when all other parents of $Y$ (acting as mediators) are held constant, and it is identifiable if and only if $P^*_{Pa(Y)}(y)$ is identifiable [16, p. 128].

The selection diagram in Figure 1(a) does not contain any Y-rooted sC-trees as subgraphs and therefore the direct effect (causal effects of Y's parents on Y) is indeed transportable. In fact, the transportability of CDE can be determined by a more visible criterion:

**Corollary 1.** *Let G be a selection diagram. Then for any node Y, the direct effect $P^*_{Pa(Y)}(y)$ is transportable if there is no S node pointing to Y.*

*Proof.* See Appendix 2. ∎

Generalizing to arbitrary effects, the following result provides a necessary condition for transportability whenever the whole graph is a sC-tree.

**Theorem 3.** *Let G be a Y-rooted sC-tree. Then the effects of any set of nodes in G on Y are not transportable.*

*Proof.* See Appendix 2. ∎

The next corollary demonstrates that sC-trees are obstacles to the transportability of $P^*_x(y)$ even when they do not involve $Y$, i.e., transportability is not a local problem – if there exists a node $W$ that is an ancestor of $Y$ but not necessarily "near" it, transportability is still prohibited (see Figure 3(b)). This fact anticipates that transporting causal effects for singletons is not necessarily easier than the general problem of transportability.

**Corollary 2.** *Let G be a selection diagram, and **X** and **Y** a set of variables. If there exists a node W that is an ancestor of some node $Y \in$ **Y** such that there exists a W-rooted sC-tree which contains any variables in **X**, then $P^*_{\mathbf{x}}(\mathbf{y})$ is not transportable.*

*Proof.* See Appendix 2. ∎

We now generalize the definition of sC-trees (and Theorem 3) in two ways: first, $Y$ is augmented to represent a set of variables; second, S-nodes can point to any variable within the sC-component, not necessarily to root nodes. For instance, consider the graph $G$ in Figure 4. Note that there is no Y-rooted sC-tree nor W-rooted sC-tree in $G$ (where $W$ is an ancestor of $Y$), and so the previous results cannot be applied even though the effect of $X$ on $Y$ is not transportable in $G$ – still, there exists a Y-rooted sC-forest in $G$, which will prevent the transportability of the causal effect.

**Definition 9** (sC-forest). *Let G be a selection diagram, where **Y** is the maximal root set. Then G is a **Y**-rooted sC-forest if G is a sC-component, all observable nodes have at most one child, and there is a selection node pointing to some vertex of G (not necessarily in **Y**).*

Building on [27], we introduce a structure that witnesses non-transportability characterized by a pair of sC-forests. Transportability will be shown impossible whenever such structure exists as an edge subgraph of the given selection diagram.

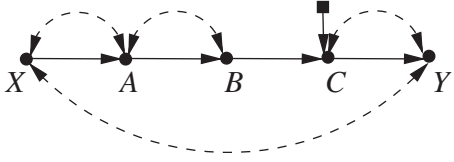

**Figure 4** Example of a selection diagram in which $P^*(y|do(x))$ is not transportable, there is no sC-tree but there is a sC-tree.

**Definition 10** (s-hedge). *Let* $\mathbf{X}, \mathbf{Y}$ *be set of variables in G. Let* $F, F'$ *be* $\mathbf{R}$*-rooted sC-forests such that* $F \cap \mathbf{X} \neq 0$, $F' \cap \mathbf{X} = 0$, $F' \subseteq F$, $\mathbf{R} \subset An(Y)_{G_{\overline{X}}}$. *Then F and F' form a s-hedge for* $P^*_\mathbf{x}(\mathbf{y})$ *in G.*

For instance, in Figure 4, the sC-forests $F' = \{C, Y\}$, and $F = F' \cup \{X, A, B\}$ form a s-hedge to $P^*_\mathbf{x}(y)$.[10] The idea here is similar to the hedge [27], and we can see a s-hedge as a growing sC-forest $F'$, which does not intersect $\mathbf{X}$, to a larger sC-forest $F$ that do intersect $\mathbf{X}$.

We state below the formal connection between s-hedges and non-transportability.

**Theorem 4.** *Assume there exist* $F, F'$ *that form a s-hedge for* $P^*_x(y)$ *in* $\Pi$ *and* $\Pi^*$. *Then* $P^*_\mathbf{x}(\mathbf{y})$ *is not transportable from* $\Pi$ *to* $\Pi^*$.

*Proof.* See Appendix 2. ∎

To prove that the s-hedges characterize non-transportability in selection diagrams, we construct in the next section an algorithm which transport any causal effects that do not contain a s-hedge.

## 5 A complete algorithm for transportability of joint effects

The algorithm proposed to solve transportability is called **sID** (see Figure 5) and extends previous analysis and algorithms of identifiability given in [13, 26, 27, 32, 34]. We choose to start with the version provided by Shpitser (called **ID**) since the hedge structure is explicitly employed, which will show to be instrumental to prove completeness. We build on two observations developed along the article:

1. Transportability: Causal relations can be partitioned into trivially and directly transportable.
2. Non-transportability: The existence of a s-hedge as an edge subgraph of the inputted selection diagram can be used to prove non-transportability.

The algorithm **sID** first applies the typical c-component decomposition on top of the inputted selection diagram D (which, by definition, is also a causal diagram of $\Pi^*$), partitioning the original problem into smaller blocks (call these blocks sc-factors) until either the entire expression is transportable or it runs into the problematic s-hedge structure.

More specifically, for each sc-factor Q, **sID** tries to directly transport Q. If it fails, **sID** tries to trivially transport Q, which is equivalent to solving an ordinary identification problem. **sID** alternates between these two types of transportability, and whenever it exhausts the possibility of applying these operations, it exits with failure with a counterexample for transportability – that is, the graph local to the faulty call witnesses the non-transportability of the causal query since it contains a s-hedge as edge subgraph.

Before showing the more formal properties of **sID**, we demonstrate how **sID** works through the transportability of $Q = P^*(y|do(x))$ in the graph in Figure 2.

---

**10** Note that, by definition, at least one S-node has to appear in both $F', F$.

1. **if** $\mathbf{x} = \emptyset$, **return** $\sum_{\mathbf{V}\setminus\mathbf{Y}} P^*(\mathbf{V})$
2. **if** $\mathbf{V}\setminus An(\mathbf{Y})_D \neq \emptyset$,
   **return** $\mathrm{sID}(\mathbf{y}, \mathbf{x} \cap An(\mathbf{Y})_D, \mathbf{Z}, \sum_{\mathbf{V}\setminus An(\mathbf{Y})_D} P^*, I, An(\mathbf{Y})_D)$
3. Set $\mathbf{W} = (\mathbf{V}\setminus\mathbf{X})\setminus An(\mathbf{Y})_{D_{\overline{\mathbf{X}}}}$.
   **if** $\mathbf{W} \neq \emptyset$, **return** $\mathrm{sID}(\mathbf{y}, \mathbf{x}\cup\mathbf{w}, \mathbf{Z}, P^*, I, D)$
4. **if** $\mathcal{C}(D\setminus\mathbf{X}) = \{C_0, C_1, ..., C_k\}$,
   **return** $\sum_{\mathbf{V}\setminus\{\mathbf{Y},\mathbf{X}\}} \prod_i \mathrm{sID}(c_i, \mathbf{V}\setminus c_i, \mathbf{Z}, P^*, I, D)$
5. **if** $\mathcal{C}(D\setminus\mathbf{X}) = \{C_0\}$
6.    **if** $\mathcal{C}(D) \neq \{D\}$,
7.     **if** $C_0 \in \mathcal{C}(D)$, **return** $\sum_{s\setminus\mathbf{Y}} \prod_{i|V_i \in S} P^*(v_i|V_D^{(i-1)})$
8.     **if** $(\exists C')C_0 \subset C' \in \mathcal{C}(D)$, **return** $\mathrm{sID}(\mathbf{y}, \mathbf{x}\cap C', C'\setminus\mathbf{X},$
   $\prod_{i|V_i \in C'} P^*(V_i|V_D^{(i-1)}\cap C', v_D^{(i-1)}\setminus C'), I, C')$.
9.    **else**,
10.     **if** $(S \perp\!\!\!\perp \mathbf{Y} \mid \mathbf{X})_{D_{\overline{\mathbf{X}}}}$, **return** $P(\mathbf{y}|do(\mathbf{x}), \mathbf{Z})$
11.     **else**, **FAIL**$(D, C_0)$

**Figure 5** Modified version of identification algorithm capable of recognizing transportable relations.

Since $D = An(Y)$ and $\mathcal{C}(D\setminus\{X\}) = (C_0, C_1, C_2)$, where $C_0 = D(\{Z\})$, $C_1 = D(\{W\})$, and $C_2 = D(\{V, Y\})$, we invoke line 4 and try to transport respectively $Q_0 = P^*_{x,w,v,y}(z)$, $Q_1 = P^*_{x,z,v,y}(w)$, and $Q_2 = P^*_{x,z,w}(v,y)$. Thus the original problem reduces to transporting $\sum_{z,w,v} P^*_{x,w,v,y}(z) P^*_{x,z,v,y}(w) P^*_{x,z,w}(v,y)$.

Evaluating the first expression, **sID** triggers line 2, noting that nodes that are not ancestors of $Z$ can be ignored. This implies that $P^*_{x,w,v,y}(z) = P^*(z)$ with induced subgraph $G_0 = \{X \to Z, X \leftarrow U_{xz} \to Z\}$, where $U_{xz}$ stands for the hidden variable between $X$ and $Z$. **sID** goes to line 5, in which in the local call $\mathcal{C}(D\setminus\{X\}) = \{G_Z\}$. In the sequel, **sID** goes to line 9 since $G_0$ contains only one sC-component. Note that in the ordinary identifiability problem the procedure would fail at this point, but **sID** proceeds to line 10 testing whether $(S \perp\!\!\!\perp \{Z\}|\{X\})_{D_{\overline{X}}}$. The test comes true, which makes **sID** directly transport $Q_0$ with data from the experimental population $\Pi$, i.e., $P^*_x(z) = P_x(z)$.

Evaluating the second expression, **sID** again triggers line 2, which implies that $P^*_{x,z,v,y}(w) = P^*_{x,z}(w)$ with induced subgraph $G_1 = \{X \to Z, Z \to W, X \leftarrow U_{xz} \to Z\}$. **sID** goes to line 5, in which in the local call $\mathcal{C}(D\setminus\{X,Z\}) = \{G_W\}$. Thus it proceeds to line 6 testing whether there are more than one sC-components. The test comes true (since $G_W \in \mathcal{C}(G_1)$), which makes **sID** to trivially transport $Q_1$ with observational data from $\Pi^*$, i.e., $P^*_{x,z}(w) = P^*(w|x,z)$.

Evaluating the third expression, **sID** goes to line 5 in which $\mathcal{C}(D\setminus\{X,Z,W\}) = \{G_2\}$, where $G_2 = \{V \to Y, S \to V, V \leftarrow U_{vy} \to Y\}$. It proceeds to line 6 testing whether there is more than one component, which is true in this case. It reaches line 8, in which $C' = G_0 \cup G_2 \cup \{X \leftarrow U_{xy} \to Y\}$. Thus it tries to transport $Q_{2'} = P^*_{x,z}(v,y)$ over the induced graph $C'$, which stands for ordinary identification, and yields (after trivial simplifications) $\sum_v P^*(v|w) P^*(y|v)$. The return of these calls composed coincide with the expression provided in the first section.

We prove next soundness and completeness of **sID**.

**Theorem 5** (soundness). *Whenever **sID** returns an expression for $P^*_\mathbf{x}(\mathbf{y})$, it is correct.*

*Proof.* See Appendix 2. ∎

**Theorem 6.** *Assume **sID** fails to transport $P^*_\mathbf{x}(\mathbf{y})$ (executes line 11). Then there exists $\mathbf{X}' \subseteq \mathbf{X}$, $\mathbf{Y}' \subseteq \mathbf{Y}$, such that the graph pair $D, C_0$ returned by the fail condition of **sID** contain as edge subgraphs sC-forests $F, F'$ that form a s-hedge for $P^*_{\mathbf{x}'}(\mathbf{y}')$.*

*Proof.* See Appendix 2. ∎

**Corollary 3** (completeness). ***sID** is complete.*

*Proof.* See Appendix 2. ∎

**Corollary 4.** $P^*_{\mathbf{x}}(\mathbf{y})$ *is transportable from* $\Pi$ *to* $\Pi^*$ *in G if and only if there is not s-hedge for* $P^*_{\mathbf{x}'}(\mathbf{y}')$ *in G for any* $\mathbf{X}' \subseteq \mathbf{X}$ *and* $\mathbf{Y}' \subseteq \mathbf{Y}$.

*Proof.* See Appendix 2. ∎

**Theorem 7.** *The rules of do-calculus, together with standard probability manipulations are complete for establishing transportability of all effects of the form* $P^*_{\mathbf{x}}(\mathbf{y})$.

*Proof.* See Appendix 2. ∎

# 6 Other perspectives on generalizability

Many problems in statistics and causal inference can be framed as problems of generalizability, though inherently different from that of transportability.

Consider, for example, classical statistical inference, it can be viewed as a generalization from properties of a *random* sample $\Pi_S$ of a population $\Pi$ to properties of the population $\Pi$ itself. Two centuries of statistical analysis have rendered this task well understood and fairly complete.

Next consider the problem of causal inference, that is, to estimate causal-effects from observational studies (given a set of causal assumptions). This class of problems can be viewed as a generalization from a population under observational regime to a population under experimental regime. Since the imposition of experimental regime (e.g., forcing individuals to receive treatment) induces a behavioral change in the population, the problem can be viewed as generalization between two diverse populations. Fortunately, the disparities between the two populations are local (assumes atomic interventions), involving only the treatment assignment mechanism and, so, with the help of model assumptions, a complete solution to the problem can be obtained (using do-calculus). We can decide algorithmically whether the assumptions at hand are sufficient for estimating a given causal effect and, if the answer is affirmative, we can derive its estimand.

An important variant in causal inference is the task of estimating causal effects from surrogate experiments, namely, experiments in which a surrogate set of variables $Z$ are manipulated, rather than the one ($X$) whose effect we seek to estimate.[11] This variant too can be viewed as an exercise in generalization, this time from a population under regime $do(Z = z)$ to that same population under regime $do(X = x)$. A complete solution to this problem is reported in [37].

Another challenge of generalizability flavor arises, in both observational and experimental studies, when samples $\Pi_S$ are not randomly drawn from the population of interest $\Pi$, but are selected preferentially, depending on the values taken by a set $V_S$ of variables. This problem, known as "selection bias" (or "sampling selection bias"), has received due attention in epidemiology, statistics, and economics [38–41] and can be viewed as a generalization from the sampled population to the population at large, when little is known about their relationships save for qualitative assumptions about the selection mechanism. Graphical models were used to improve the understanding of the problem [42–45] and gave rise to several conditions for recovering from selection bias when the probability of selection is available.

Likewise, Refs. 21, 46, 47 tackle variants of the sample selection problem assuming that certain relationships are invariant between the two groups (i.e., sample and population). The former assumed knowledge of the probability of selection in each of the principal stratum, while the latter exploited (using propensity score analysis) the availability of the probability of selection in each combination of covariates.

---

**11** A surrogate variable is different from instrumental variable in that the former should lead to the identification of causal effect even in nonparametric models; IV methods are limited to "local" causal effects (so-called LATE [48]).

More recently, Didelez et al. [49] studied conditions for recovering from selection bias when no quantitative knowledge is available about selection probabilities. Bareinboim and Pearl [50] extended these conditions and provided a complete characterization, together with an algorithm, for deciding when a bias-free estimate of the odds ratio (OR) can be recovered from selection-biased data. They also developed methods using instrumental variables that recover other effect measures when information about the target population is available for some variables (see also Ref. 51).

The problem of transportability is fundamentally different from the other problems of generalizability discussed above. Transportability deals with two distinct populations that are different both in their inherent characteristics (encoded by the S variables) and the regimes under which they are studied (i.e., experimental vs. observational).

Hernán and VanderWeele [52] addressed a problem related to transportability in the context of "compound treatments," namely, treatments that can be implemented in multiple versions (e.g., "exercise at least 15 minutes a day"). Transportability arises when we wish to predict the response of a population that implements one version of the treatment from a study on another population, in which another version is implemented. Petersen [53] showed that this problem is a variant of the general problem treated in Ref. 1, to which this article provides an algorithmic solution.

Finally, it is important to mention two recent extensions of the results reported in this article. Bareinboim and Pearl [2] have addressed the problem of transportability in cases where only a limited set of experiments can be conducted at the source environment. Subsequently, the results were generalized to the problem of "meta-transportability," that is, pooling experimental results from multiple and disparate sources to synthesize a consistent estimate of a causal relation at yet another environment, potentially different from each of the formers [3].

## 7 Conclusions

Informal discussions concerning the difficulties of generalizing experimental results across populations have been going on for almost half a century [4, 5, 54–56] and appear to accompany every textbook in experimental design. By and large, these discussions have led to the obvious conclusions that researchers should be extremely cautious about unwarranted generalization, that many threats may await the unwary, and that extrapolation across studies requires "some understanding of the reasons for the differences" [54, p. 11].

The formalization offered in this article embeds this discussion in a precise mathematical language and provides researchers with theoretical guarantees that, if certain conditions can be ascertained, generalization across populations can be accomplished, protected from the threats and dangers that the informal literature has accumulated.

Given judgmental assessments of how target populations may differ from those under study, the article offers a formal representational language for making these assessments precise (Definition 3) and, subsequently, deciding whether, and how, causal relations in the target population can be inferred from those obtained in experimental studies. Corollary 4 in this article provides a complete (necessary and sufficient) graphical condition for deciding this question and, whenever satisfied, we further provide an algorithm for computing the correct transport formula (Figure 5). The transport formula specifies the proper way of modifying the experimental results so as to account for differences in the populations. These transport formulae enable the investigator to select the essential measurements in both the experimental and observational studies and combine them into a bias-free estimand of the target quantity.

While the results of this article concern the transfer of causal information from experimental to observational studies, the method can also benefit in transporting statistical findings from one observational study to another [57]. The rationale for such transfer is twofold. First, information from the first study may enable researchers to avoid repeated measurement of certain variables in the target population.

Second, by pooling data from both populations, we increase the precision in which their commonalities are estimated and, indirectly, also increase the precision by which the target relationship is transported. Substantial reduction in sampling variability can be thus achieved through this decomposition [58].

Of course, our analysis is based on the assumption that the analyst is in possession of sufficient background knowledge to determine, at least qualitatively, *where* two populations may differ from one another. In practice, such knowledge may only be partially available. Still, as in every mathematical exercise, the benefit of the analysis lies primarily in understanding what must be assumed about reality for generalization to be valid, what knowledge is needed for a given task to succeed, and how sensitive conclusions are to knowledge that we do not possess.

**Acknowledgment:** A preliminary version of this article was presented at the 26th AAAI Conference, Toronto, CA, July, 2012 [59]. We appreciate the insightful comments provided by two anonymous referees. This article benefited from discussions with Onyebuchi Arah, Stuart Baker, Susan Ellenberg, Eleazar Eskin, Constantine Frangakis, Sander Greenland, David Heckerman, James Heckman, Michael Hoefler, Marshall Joffe, Rosa Matzkin, Geert Molengergh, William Shadish, Ian Shrier, Dylan Small, Corwin Zigler, and Song-Chun Zhu.

This research was supported in parts by grants from NSF #IIS-1249822, and ONR #N00014–13–1-0153 and #N00014–10–1-0933.

# Appendix 1: causal assumptions in nonparametric models

The tools presented in this article were developed in the framework of nonparametric SCM, which subsumes and unifies many approaches to causal inference.[12]

A SCM $M$ conveys a set of assumptions about how the world operates. This contrasts the statistical tradition in which a model is defined as a set of distributions (see footnote 15). Causal models is better viewed as a set of assumptions about Nature, with the understanding that each assumption (i.e., that the set of arguments of $f_i$ does not include variable $V_j$) constrains the set of distributions (like $P(v)$) that the model can generate.

The formal structure of SCM's was defined in Section 3, here we illustrate their power as inference engines. Consider a simple SCM model depicted in Figure 6(a), which represents the following three functions:

$$z = f_Z(u_Z)$$
$$x = f_X(z, u_X) \quad [10]$$
$$y = f_Y(x, u_Y),$$

where in this particular example, $U_Z$, $U_X$, and $U_Y$ are assumed to be jointly independent but otherwise arbitrarily distributed. Each of these functions represents a causal process (or mechanism) that determines

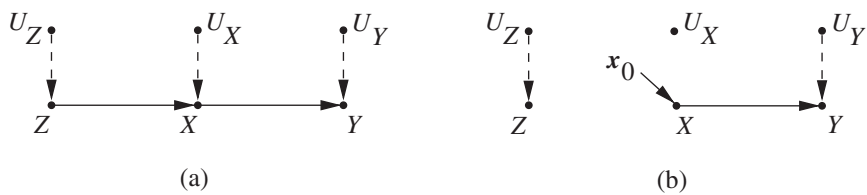

**Figure 6** The diagrams associated with (a) the structural model of eq. [6] and (b) the modified model of eq. [11], representing the intervention $do(X = x_0)$.

---

**12** We use the acronym SCM for both parametric and non-parametric representations (which is also called Structural Equation Model (SEM)), though historically, SEM practitioners preferred the parametric representation and often confuse with regression equations [60].

the value of the left variable (output) from the values on the right variables (inputs) and is assumed to be invariant unless explicitly intervened on. The absence of a variable from the right-hand side of an equation encodes the assumption that nature ignores that variable in the process of determining the value of the output variable. For example, the absence of variable $Z$ from the arguments of $f_Y$ conveys the empirical claim that variations in $Z$ will leave Y unchanged, as long as variables $U_Y$ and $X$ remain constant.

## Representing Interventions, counterfactuals, and causal effects

This feature of invariance permits us to derive powerful claims about causal effects and counterfactuals, even in nonparametric models, where all functions and distributions remain unknown. This is done through a mathematical operator called $do(x)$, which simulates physical interventions by deleting certain functions from the model, replacing them with a constant $X = x$, while keeping the rest of the model unchanged [61–63]. For example, to emulate an intervention $do(x_0)$ that holds $X$ constant (at $X = x_0$) in model $M$ of Figure 6(a), we replace the equation for $x$ in eq. [10] with $x = x_0$, and obtain a new model, $M_{x_0}$,

$$\begin{aligned} z &= f_Z(u_Z) \\ x &= x_0 \\ y &= f_Y(x, u_Y), \end{aligned} \qquad [11]$$

the graphical description of which is shown in Figure 6(b).

The joint distribution associated with the modified model, denoted $P(z, y|do(x_0))$ describes the post-intervention distribution of variables $Y$ and $Z$ (also called "controlled" or "experimental" distribution), to be distinguished from the preintervention distribution, $P(x, y, z)$, associated with the original model of eq. [10]. For example, if $X$ represents a treatment variable, $Y$ a response variable, and $Z$ some covariate that affects the amount of treatment received, then the distribution $P(z, y|do(x_0))$ gives the proportion of individuals that would attain response level $Y = y$ and covariate level $Z = z$ under the hypothetical situation in which treatment $X = x_0$ is administered uniformly to the population.[13]

In general, we can formally define the postintervention distribution by the equation

$$P_M(y|do(x)) = P_{M_x}(y) \qquad [12]$$

In words, in the framework of model $M$, the postintervention distribution of outcome $Y$ is defined as the probability that model $M_x$ assigns to each outcome level $Y = y$. From this distribution, which is readily computed from any fully specified model $M$, we are able to assess treatment efficacy by comparing aspects of this distribution at different levels of $x_0$.[14]

## Identification, d-separation and causal calculus

A central question in causal analysis is the question of *identification* in partially specified models: Given assumptions set $A$ (as embodied in the model), can the controlled (postintervention) distribution, $P(y|do(x))$, be estimated from data governed by the preintervention distribution $P(z, x, y)$?

In linear parametric settings, the question of identification reduces to asking whether some model parameter, $\beta$, has a unique solution in terms of the parameters of $P$ (say the population covariance matrix).

---

**13** Equivalently, $P(z, y|do(x_0))$ can be interpreted as the joint probability of $(Z = x, Y = y)$ under a randomized experiment among units receiving treatment level $X = x_0$. Readers versed in potential-outcome notations may interpret $P(y|do(x), z)$ as the probability $P(Y_x = y|Z_x = z)$, where $Y_x$ is the potential outcome under treatment $X = x$.
**14** Counterfactuals are defined similarly through the equation $Y_x(u) = Y_{M_x}(u)$ (see [16, Ch. 7]), but will not be needed for the discussions in this article.

In the nonparametric formulation, the notion of "has a unique solution" does not directly apply since quantities such as $Q(M) = P(y|do(x))$ have no parametric signature and are defined procedurally by simulating an intervention in a causal model $M$, as in eq. [11]. The following definition captures the requirement that $Q$ be estimable from the data:

**Definition 11** (Identifiability).[15] *A causal query $Q(M)$ is identifiable, given a set of assumptions A, if for any two models (fully specified) $M_1$ and $M_2$ that satisfy A, we have*

$$P(M_1) = P(M_2) \Rightarrow Q(M_1) = Q(M_2) \quad [13]$$

In words, the functional details of $M_1$ and $M_2$ do not matter; what matters is that the assumptions in $A$ (e.g., those encoded in the diagram) would constrain the variability of those details in such a way that equality of $P$'s would entail equality of $Q$'s. When this happens, $Q$ depends on $P$ only, and should therefore be expressible in terms of the parameters of $P$.

When a query $Q$ is given in the form of a do-expression, for example $Q = P(y|do(x), z)$, its identifiability can be decided systematically using an algebraic procedure known as the do-calculus [13]. It consists of three inference rules that permit us to map interventional and observational distributions whenever certain conditions hold in the causal diagram G.

The conditions that permit the application these inference rules can be read off the diagrams using a graphical criterion known as d-separation [65].

**Definition 12** (d-separation). *A set S of nodes is said to block a path p if either*

1. *p contains at least one arrow-emitting node that is in S, or*
2. *p contains at least one collision node that is outside S and has no descendant in S.*

*If S blocks all paths from set X to set Y, it is said to "d-separate X and Y," and then, it can be shown that variables X and Y are independent given S, written $X \perp\!\!\!\perp Y | S$.*[16]

D-separation reflects conditional independencies that hold in any distribution $P(v)$ that is compatible with the causal assumptions $A$ embedded in the diagram. To illustrate, the path $U_Z \to Z \to X \to Y$ in Figure 6(a) is blocked by $S = \{Z\}$ and by $S = \{X\}$, since each emits an arrow along that path. Consequently we can infer that the conditional independencies $U_Z \perp\!\!\!\perp Y | Z$ and $U_Z \perp\!\!\!\perp Y | X$ will be satisfied in any probability function that this model can generate, regardless of how we parametrize the arrows. Likewise, the path $U_Z \to Z \to X \leftarrow U_X$ is blocked by the null set $\{\emptyset\}$, but it is not blocked by $S = \{Y\}$ since $Y$ is a descendant of the collision node $X$. Consequently, the marginal independence $U_Z \perp\!\!\!\perp U_X$ will hold in the distribution, but $U_Z \perp\!\!\!\perp U_X | Y$ may or may not hold.[17]

## The rules of *do*-calculus

Let $X$, $Y$, $Z$, and $W$ be arbitrary disjoint sets of nodes in a causal DAG $G$. We denote by $G_{\overline{X}}$ the graph obtained by deleting from $G$ all arrows pointing to nodes in $X$. Likewise, we denote by $G_{\underline{X}}$ the graph obtained by deleting from $G$ all arrows emerging from nodes in $X$. To represent the deletion of both incoming and outgoing arrows, we use the notation $G_{\overline{XZ}}$.

---

[15] This definition appears to be similar to, but differ fundamentally from the standard statistical definition [64, p. 22] which deals with the unidentifiability of the parameter set $\theta$ from a distribution $P_\theta$. In our case, the query $Q = P(Y|do(x))$ is not a parameter of $P$ (see [22, p. 77]).
[16] See Hayduk et al. [66], Glymour and Greenland [67], and Pearl [16, p. 335] for a gentle introduction to *d*-separation.
[17] This special handling of collision nodes (or *colliders*, e.g., $Z \to X \leftarrow U_x$) reflects a general phenomenon known as *Berkson's paradox* [68], whereby observations on a common consequence of two independent causes render those causes dependent. For example, the outcomes of two independent coins are rendered dependent by the testimony that at least one of them is a tail.

The following three rules are valid for every interventional distribution compatible with $G$.

**Rule 1** (Insertion/deletion of observations):

$$P(y|do(x), z, w) = P(y|do(x), w) \text{ if } (Y \perp\!\!\!\perp Z | X, W)_{G_{\overline{X}}} \qquad [14]$$

**Rule 2** (Action/observation exchange):

$$P(y|do(x), do(z), w) = P(y|do(x), z, w) \text{ if } (Y \perp\!\!\!\perp Z | X, W)_{G_{\overline{X}\underline{Z}}} \qquad [15]$$

**Rule 3** (Insertion/deletion of actions):

$$P(y|do(x), do(z), w) = P(y|do(x), w) \text{ if } (Y \perp\!\!\!\perp Z | X, W)_{G_{\overline{XZ(W)}}}, \qquad [16]$$

where $Z(W)$ is the set of $Z$-nodes that are not ancestors of any $W$-node in $G_{\overline{X}}$.

To establish identifiability of a query $Q$, one needs to repeatedly apply the rules of do-calculus to $Q$, until the final expression no longer contains a do-operator[18]; this renders it estimable from non-experimental data. The do-calculus was proven to be complete to the identifiability of causal effects in the form $Q = P(y|do(x), z)$ [69, 70], which means that if $Q$ cannot be expressed in terms of the probability of observables $P$ by repeated application of these three rules, such an expression does not exist.

We shall see that, to establish transportability, the goal will be different; instead of eliminating do-operators, we will need to separate them from a set of variables S that represent disparities between populations.

## Appendix 2

**Theorem 2.** *Let $G$ be a selection diagram. Then for any node $Y$, the direct effect $P^*_{Pa(Y)}(y)$ is transportable if there is no subgraph of $G$ which forms a $Y$-rooted sC-tree.*

*Proof.* We known from Tian [71, Theorem 22] that whenever there exists no subgraph $G_T$ of $G$ satisfying all of the following: (i) $Y \in T$; (ii) $G_T$ has only one c-component, $T$ itself; (iii) All variables in $T$ are ancestors of $Y$ in $G_T$, the direct effect on $Y$ is identifiable, as sC-trees are structures of this type. Further Shpitser and Pearl [27, Theorem 2] showed that the same holds for C-trees, which also implies the inexistence of a sC-trees. Since such structure does not show up in $G$, the target quantity is identifiable, and hence transportable.

It remains to show that the same holds whenever there exists a subgraph that is a C-tree and in which no S node points to Y, i.e., there is no Y-rooted sC-tree at all. It is true that $(S \perp\!\!\!\perp Y | Pa(Y))_{G_{\overline{Pa(Y)}}}$, given that all directed paths from $S$ to $Y$ are closed. This follows from the following facts: (1) all paths from $S$ passing through $Y$'s ancestors were cut in $G_{\overline{Pa(Y)}}$; (2) all bidirected paths were also closed given that the conditioning set contains only root nodes, and a connection from $S$ must pass through at least one collider; (3) transportability does not depend on descendants of Y (by argument similar to Tian [71, Lemma 9]). Thus, it follows that we can write $P^*_{Pa(Y)}(Y) = P_{Pa(Y)}(Y|S) = P_{Pa(Y)}(Y)$, concluding the proof. ∎

**Corollary 1.** *Let $G$ be a selection diagram. Then for any node $Y$, the direct effect $P^*_{Pa(Y)}(y)$ is transportable if there is no S node pointing to Y.*

*Proof.* Follows directly from Theorem 2. ∎

---

**18** Such derivations are illustrated in graphical details in Ref. [16, p. 87].

**Lemma 5.** *The exclusive OR (XOR) function is commutative and associative.*

*Proof.* Follows directly from the definition of the XOR function. ∎

**Remark 1.** The construction given below is a strict generalization of Theorem 1, and it is useful because it will provide a simplified construction of the one provided in Theorem 1, and also set the tone for proofs of generic graph structures which will in the sequel show to be instrumental in proving non-transportability in arbitrary structures.

**Theorem 3.** *Let G be a Y-rooted sC-tree. Then the effects of any set of nodes in G on Y are not transportable.*

*Proof.* The proof will proceed by constructing a family of counterexamples. For any such $G$ and any set $\mathbf{X}$, we will construct two causal models $M_1$ and $M_2$ that will agree on $\langle P, P^*, I \rangle$, but disagree on the interventional distribution $P_x^*(y)$.

Let the two models $M_1$, $M_2$ agree on the following features. All variables in $\mathbf{U} \cup \mathbf{V}$ are binary. All exogenous variables are distributed uniformly. All endogenous variables except $Y$ are set to the bit parity (sum mod 2) of the values of their parents. The two models differ in respect to $Y$'s definition. Consider the function for $Y$, $f_Y : U, Pa(Y) \to Y$ to be defined as follows:

$$\begin{cases} M_1 : Y = ((pa(Y) \otimes u) \otimes s) \\ M_2 : Y = ((pa(Y) \otimes u) \vee s) \end{cases}$$

**Lemma 6.** *The two models agree in the distributions $\langle P, P^*, I \rangle$.*

*Proof.* Since the two models agree on $P(U)$ and all functions except $f_Y$, it suffices to show that $f_Y$ maintains the same input/output behavior in both models for each domains.

**Subclaim 1**: Let us show that both models agree in the observational and interventional distributions relative to domain $\Pi$, i.e., the pair $\langle P, I \rangle$. The index variable $S$ is set to 0 in $\Pi$, and $f_Y$ evaluates to $(pa(Y) \otimes u)$ in both models, which proves the subclaim.

**Subclaim 2**: Let us show that both models agree in the observational distribution relative to $\Pi^*$, i.e., $P^*$. The index variable $S$ is set 1 in $\Pi^*$, and $f_Y$ evaluates to $((pa(Y) \otimes u) \otimes 1)$ in $M_1$, and 1 in $M_2$. Since the evaluation in $M_1$ can be rewritten as $\neg((pa(Y) \otimes u))$, it remains to show that $(pa(Y) \otimes u)$ always evaluates to 0.

This fact is certainly true, consider the following observations: a) each variable in $U$ has exactly two endogenous children; b) the given tree has $Y$ as the root; c) all functions are $XOR$ – these imply that $Y$ is computing the bit parity of the sum of all $U$ nodes, which turns out to be even, and so evaluates to 0 and proves the subclaim. ∎

**Lemma 7.** *For any set $\mathbf{X}$, $P_1(Y|do(\mathbf{X}), S = 1) \neq P_2(Y|do(\mathbf{X}), S = 1)$.*

*Proof.* Given the functional description and the discussion in the previous Lemma, the function $f_Y$ evaluates always to 1 in $M_2$.

Now let us consider $M_1$. Note that performing the intervention and cutting the edges going toward $\mathbf{X}$ creates an asymmetry on the sum of the bidirected edges departing from $\mathbf{U}$, and consequently in the sum performed by $Y$. It will be the case that some $\mathbf{U}'$ will appear only once in the expression of $Y$. Therefore, depending on the assignment $\mathbf{X} = \mathbf{x}$, we will need to evaluate the sum (mod 2) over $\mathbf{U}'$ in $Y$ or its negation, which given the uniformity of the distribution of $\mathbf{U}$ will yield $P_1(Y|do(\mathbf{X}), S = 1) = 1/2$ in both cases. ∎

By Lemma 2, Lemmas 6 and 7 together prove Theorem 3. ∎

**Corollary 2.** *Let G be a selection diagram, let* **X** *and* **Y** *be set of variables. If there exists a node W which is an ancestor of some node $Y \in$ **Y** and such that there exists a W-rooted sC-tree which contains any variables in* **X**, *then $P^*_\mathbf{x}(\mathbf{y})$ is not transportable.*

*Proof.* Fix a $W$-rooted sC-tree T, and a path $p$ from $W$ to $Y$. Consider the graph $p \cup T$. Note that in this graph $P^*_\mathbf{x}(Y) = \sum_w P^*_\mathbf{x}(w) P^*(Y|w)$. From the last Theorem $P^*_\mathbf{x}(w)$ is not transportable, it is now easy to construct $P^*(Y|W)$ in such a way that the mapping from $P_\mathbf{x}(W)$ to $P_\mathbf{x}(Y)$ is one to one, while making sure all distributions are positive.

**Remark 2.** The previous results comprised cases in which there exist sC-trees involved in the non-transportability of $Y$ – i.e., $Y$ or some of its ancestors were roots of a given sC-tree. In the problem of identifiability, the counterpart of sC-trees (i.e., C-trees) suffices to characterize non-identifiability for singleton $Y$. But transportability is more subtle and this is not the case here – it not only depends on $X$ and $Y$ "locations" in the graph, but also the relative position of the S-nodes. Consider Figures 4 and 7(a) (called sp-graph). In these graphs there is no sC-tree but the effect of $X$ on $Y$ is still non-transportable.

The main technical subtlety here is that in sC-trees, a S-node combines its effect with a X-node intersecting in the root node (considering only the bidirected edges), which is not the case for non-transportability in general. Note that in the graphs in Figure 4, and the sp-graph, the nodes S and X intersect first through ordinary edges and meet through bidirected edges only on the Y node. This implies a certain "asynchrony" because, in the structural sense, the existence of a S-node implies a difference in the structural equations between domains, but only this difference does not imply non-transportability (for instance, $P^*_x(z)$ is transportable in the sp-graph even though the equations of $Z$ being different in both models).

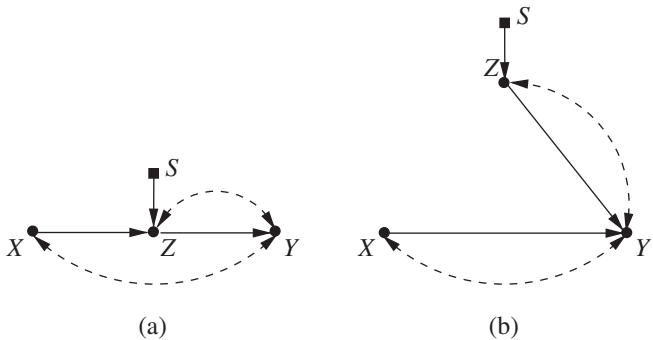

**Figure 7** Selection diagrams in which $P^*(y|do(x))$ is not transportable, there is no sC-tree but there is a sC-forest. These diagrams will be used as basis for the general case; the first diagram is named sp-graph and the second one sb-graph.

The key idea to produce a proof for non-transportability in these cases is to keep the effect of S-nodes after intersecting with X "dormant" until they reach the target Y and then manifest. We implement this idea in the next two proofs, which can be seen as base cases, and should pavement the way for the most general problem.

**Theorem 8.** $P^*_x(y)$ *is not transportable in the sp-graph (Figure 7(a)).*

*Proof.* We will construct two causal models $M_1$ and $M_2$ compatible with the sp-graph that will agree on $\langle P, P^*, I \rangle$, but disagree on the interventional distribution $P^*_x(y)$.

Let us assume that all variables in $U \cup V$ are binary, and let $U_1$ be the common cause of $X$ and $Y$, $U_2$ be the common cause of $Z$ and $Y$, and $U_3$ be the random disturbance exclusive to $Z$. Let $M_1$ and $M_2$ be defined as follows:

$$M_1 = \begin{cases} X = U_1 \\ Z = (((X \otimes U_2 \otimes 1) \otimes U_3) \vee S) \otimes (S \wedge (X \otimes U_2)) \\ Y = Z \otimes U_1 \otimes U_2 \end{cases}$$

and:

$$M_2 = \begin{cases} X = U_1 \\ Z = (((U_2 \otimes 1) \otimes U_3) \vee S) \otimes (S \wedge U_2) \\ Y = Z \otimes U_2 \end{cases}$$

Both models agree in respect to $P(\mathbf{U})$, which is defined as follows: $P(U_1) = P(U_2) = P(U_3) = 1/2$.

**Lemma 8**. *The two models agree in the distributions $\langle P, P^*, I \rangle$.*

*Proof.* **Subclaim 1**: Let us show that both models agree in the observational and interventional distributions relative to domain $\Pi$, i.e., the pair $\langle P, I \rangle$. In both models $X$ has the same expression, which entails the same (uniform) probabilistic behavior in both cases. The index variable $S$ is set to 0 in $\Pi$, and $Z$ evaluates to $(X \otimes U_2 \otimes 1 \otimes U_3)$ in $M_1$ and $(U_2 \otimes 1 \otimes U_3)$ in $M_2$. Clearly, for any value of $X = x$, since $U$ is the same and uniformly distributed in both models, we obtain the same (uniform) input/output probabilistic behavior in $M_1$ and $M_2$ (note that $U_2, U_3$ can freely vary independently of $X$). In similar way, $Y$ evaluates to $(1 + U_3)$ in both models, which entails the same (uniform) input/output probabilistic behavior in both models. In regard to $do(X = x)$, it is clear that $Z$ did not depend (probabilistically) on the specific value of $X$, and so the equality between both models follows. For the case when we have $do(Z = z)$, $Y$ evaluates to $(Z \otimes U_1 \otimes U_2)$ in $M_1$ and $(Z \otimes U_2)$ in $M_2$, and given the uniformity of $U$, they preserve the same (uniform) input/output probabilistic behavior. (For a more elaborated argument, see Theorem 4 below.)

**Subclaim 2**: Let us show that both models agree in the observational distribution $P^*$ relative to $\Pi^*$. The index variable $S$ is set 1 in $\Pi^*$, $f_Z$ evaluates to $(X \otimes U_2 \otimes 1)$ in $M_1$, and $(U_2 \otimes 1)$ in $M_2$. Again, for any value of $X$, together with the uniformity of $U$, we obtain the same (uniform) input/output probabilistic behavior in both models (note again that $U_2$ can freely vary independently of variations of $X$, and so $Z$). Further, $f_Y$ evaluates to 1 in both models, which yields the same (uniform) input/output behavior in both models. (To guarantee positivity, we can apply the trick of making a new $f_{Y'}()$ such that $f_{Y'}()$ returns 0 half the time, and $f_Y$ the other half (i.e., set $f_{Y'}() = [f_Y() \wedge C]$, where $C$ is a fair coin.) ∎

**Lemma 9**. *There exist values of such that $X, Y$ $P_1(Y|do(X), S = 1) \neq P_2(Y|do(X), S = 1)$.*

*Proof.* Fix $X = 1, Y = 1$. First notice that $f_Z$ evaluates to $U_2$ in $M_1$ and $(U_2 \otimes 1)$ in $M_2$. Given that $U_2$ is uniformly distributed, both quantities coincide (and they represent the effect of $X$ on $Z$, which is transportable in $G$). Now the evaluation of $f_Y$ in $M_1$ reduces to $U_1$, while it reduces to 1 in $M_2$, which show disagreement and finishes the proof of this Lemma. ∎

By Lemma 2, Lemmas 8 and 9 together prove Theorem 8. ∎

**Remark 3.** There exists a different sort of asymmetry in the case of Figure 7(b) (called *sb*-graph), and the nodes $X$ and $S$ do not intersect before meeting $Y$ – i.e., they have disjoint paths and $Y$ lies precisely in their intersection.

Still, this case is not the same of having a *sC*-tree because in *sb*-graphs we need to keep the equality from the $S$ nodes to $Y$ until $S$ intersects $X$ on $Y$. Employing a similar construct as in the *sp*-graph, we keep the effect of $S$ dormant until it reaches $Y$ and then emerges.

**Theorem 9.** *$P_x^*(y)$ is not transportable in the sb-graph (Figure 7(b)).*

*Proof.* We construct two causal models $M_1$ and $M_2$ compatible with the *sb*-graph that will agree on $\langle P, P^*, I \rangle$, but disagree on the interventional distribution $P_x^*(y)$.

Let us assume that all variables in $\mathbf{U} \cup \mathbf{V}$ are binary, and let $U_1$ be the common cause of $X$ and $Y$, $U_2$ be the common cause of $Z$ and $Y$, and $U_3$ be the random disturbance exclusive to $X$. Let $M_1$ and $M_2$ agree with the following definitions:

$$M_1, M_2 = \begin{cases} X = U_1 \\ Z = ((U_3 \otimes U_2 \otimes 1) \vee S) \otimes (S \wedge U_2)) \end{cases}$$

and disagree in respect to $Z$ as follows:

$$\begin{cases} M_1 : Y = Z \otimes U_2 \\ M_2 : Y = X \otimes Z \otimes U_1 \otimes U_2 \end{cases}$$

Both models also agree in respect to $P(\mathbf{U})$, which is defined as follows:

$$P(U_1) = P(U_2) = P(U_3) = 1/2$$

**Lemma 10.** *The two models agree in the distributions $\langle P, P^*, I \rangle$.*

*Proof.* **Subclaim 1:** Let us show that both models agree in the observational and interventional distributions relative to domain $\Pi$, i.e., the pair $\langle P, I \rangle$. The index variable $S$ is set to 0 in $\Pi$, and $\{X, Z\}$ are defined in the same way in both models, and so it suffices to analyze $Y$, which in this case evaluates to $(U_3 \otimes 1)$ in both models, preserving the same (uniform) probabilistic behavior. Given that, it is not difficult to see that both models also evaluate in the same way when considering the interventions in $I$.

**Subclaim 2:** Let us show that both models agree in the observational distribution $P^*$ relative to $\Pi^*$. The index variable $S$ is set 1 in $\Pi^*$, given that $\{X, Z\}$ are defined in the same way in both models, together with the uniformity of $U$ make them evaluate in the same way in both models, and $Y$ evaluates to 1 in both models. (As in Lemma 8, the same trick to make the distribution positive could be applied here.) ∎

**Lemma 11.** *There exist values of $X, Y$ such that $P_1(Y|do(X), S = 1) \neq P_2(Y|do(X), S = 1)$.*

*Proof.* Fix $X = 1, Y = 1$. First notice that $f_Z$ evaluates to $(U_2 \otimes 1)$ in both models, and the evaluation of $f_Y$ in $M_1$ reduces to 1, while it reduces to $U_1$ in $M_2$. It follows that in $M_1$, $f_Y$ evaluates to 1 with probability 1, while in $M_2$ it evaluates to 1 with probability $P(U_1 = 1)$, which disagree by construction, finishing the proof of this Lemma. ∎

By Lemma 2, Lemmas 10 and 11 together prove Theorem 9. ∎

**Remark 4.** There are two complementary components to forge a general scheme to prove arbitrary non-transportability. First, the construct of Theorem 4 shows how to prove non-transportability for general structures such as *sC*-trees. In the sequel, the specific proofs of non-transportability for the sp-*graph* (Theorem 9) and sb-*graph* (Theorem 10) partition the possible interactions between $X$, $S$ and $Y$. In the former, $X$ and $S$ intersect before meeting with $Y$, while in the latter they have disjoint paths and $Y$ lies in their intersection. In the sequel, the proof for the general case combines these analyses, which we show below.

**Theorem 4.** *Assume there exist $F, F'$ that form a s-hedge for $P^*_\mathbf{x}(\mathbf{y})$ in $\Pi$ and $\Pi^*$. Then $P^*_\mathbf{x}(\mathbf{y})$ is not transportable from $\Pi$ to $\Pi^*$.*

*Proof.* We first consider counterexamples with the induced graph $H = De(F)_G \cap An(\mathbf{Y})_{G_{\overline{\mathbf{X}}}}$, and assume, without loss of generality, that $H$ is a forest. We construct two causal models $M_1$ and $M_2$ that will agree on $\langle P, P^*, I \rangle$, but disagree on the interventional distribution $P^*_\mathbf{x}(\mathbf{y})$.

Let $F$ be an **R**-rooted *sC*-forest, let $\mathbf{V}'$ be the set of observable variables and $\mathbf{U}'$ be the set of unobservable variables in $F$. Let us assume that all variables in $\mathbf{U}' \cup \mathbf{V}'$ are binary. Call $\mathbf{W}$ the set of variables pointed by $S$-nodes in $F'$, which by the definition of *sC*-forest is guaranteed to be non-empty.

In model 1, let each $V_i \in \mathbf{V}' \setminus \mathbf{W}$ compute the bit parity of all its observable and unobservable parents (i.e., $f_i^{(1)} = \otimes (\bigcup_{V_j \in \mathbf{Pa_i}} V_j)$, where the xor is applied for each element of the set and the result computed so far), while in model 2, let $V_i$ compute the bit parity of all its parents except that any node in $F'$ disregards the parents values if the parent is in $F$ (i.e., $f_i^{(2)} = \otimes(\bigcup_{V_j \in Pa_i \cap F'} V_j)$ if $V_i$ is in $F'$, and $f_i^{(2)} = f_i^{(1)}$, otherwise).

Define $W \in W$ as follows:

$$\begin{cases} M_1 : W = ((f_w^{(1)} \otimes U_w^*) \vee S) \otimes (S \wedge (1 \otimes f_w^{(1)})) \\ M_2 : W = ((f_w^{(2)} \otimes U_w^*) \vee S) \otimes (S \wedge (1 \otimes f_w^{(2)})) \end{cases}.$$

where $f_w$ is constructed in similar way as $f_i$ in $M_1$ and $M_2$ above, and $U_w^*$ is an additional fair coin exclusively pointing to $W$. Let us call $\mathbf{U_w}$ the collection of such coins. Furthermore, let us assume that each $U_i \in \{\mathbf{U}' \setminus \mathbf{U_w}\}$ is also a fair coin (i.e., $P(U_i) = 1/2$).

**Lemma 12.** *The two models agree in the distribution of $P^*$ and there exists a value assignment $\mathbf{x}$ for $\mathbf{X}$ such that $P_1(\mathbf{Y}|do(\mathbf{x}), S = 1) \neq P_2(\mathbf{Y}|do(\mathbf{x}), S = 1)$.*

*Proof.* For $S = 1$, the result follows directly since the systems of equations in both models reduce to the construction given in Theorem 4 at [27]. ∎

**Lemma 13.** *The two models agree in the distributions $\langle P, I \rangle$.*

*Proof.* Let us show that both models agree in the observational distribution $P$ relative to domain $\Pi$. The selection variable $S$ is set to 0 in $\Pi$, and note that both systems are the same as in $\Pi^*$ except that now each variable $W \in \mathbf{W}$ has an extra variable $U_w^*$ pointing to it that should be taken into account in $W$'s evaluation, and in turn in the whole system.

We have a forest over the endogenous nodes and all functions compute the bit parity of the value of their parents, and so we can view each node as computing the sum mod 2 of its exogenous ancestors in $H$. We want to show that the distribution of each family is equally likely for each possible assignment (i.e., $P(v_i|\mathbf{pa_i}) = 1/2$, for all $v_i, \mathbf{pa_i}$).

Let us partition the analysis in two cases. First consider the case of $V_i \in \mathbf{R}$ in which there exists a $S$-node in the respective sC-tree. Note that the evaluation of $V_i$ relies only on the value of $U_w^* \in \mathbf{U_w}$ in its respective tree since $U \in \{\mathbf{U}' \setminus \mathbf{U_w}\}$ has an even number of endogenous children in $F$, and it is counted twice, so evaluates to zero (i.e., it does not affect $V_i$'s evaluation). For now, let us assume that there is only one $U_w^*$ that affects the evaluation of $V_i$. Given the uniformity of $U_w^*$, it suffices to show that $U_w^*$ can vary independently for any configuration of the parents of $V_i$.

For any configuration of $\mathbf{U}' = (U_1 = u_1, ..., U_w^* = u_w^*, ...)$, consider the corresponding evaluation of $\mathbf{Pa_i} = \mathbf{pa_i}$, and also $V_i = u_w^*$. We want to show that it is possible to flip the current value of $U_w^*$ from $u_w^*$ to $\neg u_w^*$ while preserving the parents' evaluation $\mathbf{pa_i}$. Assume this is not so. This implies that the evaluation of $Pa_i$ and $V_i$ count the same $\mathbf{U}$'s, contradiction.

To see why, consider $\mathbf{Pa_i}^* \subseteq \mathbf{Pa_i}$ the set of parents of $V_i$ that are descendents of $U_w^*$. Now, for each of these parents flip the minimum number of variables from $\mathbf{U} \setminus \mathbf{U_w}$, and call this set $\mathbf{U}^*$. (Note that this is always possible since we need at most one $U$ for each parent, which should exist by construction of sC-forest.) Now, make $U_w^* = \neg u_w^*$, and note that $\mathbf{Pa_i} = \mathbf{pa_i}$ since flipping the values of $\mathbf{U}^*$ compensates the flip of $U_w^*$. But it is also true now that $V_i$ evaluates to $\neg u_w^*$ since, in the same way as before, all other variables in $\{\mathbf{U} \setminus \mathbf{U_w}\}$ are cancelled out in $V_i$'s evaluation, including the ones in $\mathbf{U}^*$. This proves the claim.

Consider the following two facts: **Subclaim 1**: Let X and Y be two binary variables such that $P(X = x) = p \neq 1/2$ and $P(Y = y) = q = 1/2$. Then the probabilistic input/output behavior of $Z = XOR(X, Y)$ is the same of Y. The variable $Z = 1$ whenever $\{(X = 1, Y = 0), (X = 0, Y = 1)\}$, which happens with probability $pq + (1 - p)(1 - q)$. Since $q = 1/2$, the expression reduces to $p * 1/2 + (1 - p) * 1/2 = 1/2$.

**Subclaim 2**: Let X and Y be two binary variables such that $P(X = x) = P(Y = y) = p = 1/2$. Then the probabilistic input/output behavior of $Z = XOR(X, Y)$ is the same of X (or Y). This follows directly from Subclaim 1. It is clear that if there are multiple nodes from $\mathbf{U_w}$ in the evaluation of $V_i$, the same construction is also valid given the subclaim above. It is also not difficult to generalize this argument to consider root set that are not singleton, including roots in which there are not S-nodes as ancestors.

Finally, let us consider the case of $V_i \in \{F \backslash \mathbf{R}\}$. It suffices to show that the function from $\mathbf{U'} \backslash \mathbf{U_w}$ to $\mathbf{V'} \backslash \mathbf{R}$ is 1–1 when we fix $\mathbf{U_w} = \mathbf{u_w}$. We use the same argument as Shpitser. Assume this is not so, and fix two instantiations of $\mathbf{U'} \backslash \mathbf{U_w}$ that map to the same value of $\mathbf{V'} \backslash \mathbf{R}$, and differ by the set $\mathbf{U}^* = \{U_1, ..., U_k\}$. Since the bidirected edges form a spanning tree, there exists $\mathbf{V}^*$ with an odd number of parents in $\mathbf{U}^*$ (and were not in $\mathbf{R}$, by construction). Order them topologically and let the topmost be called $X$. Note that if we flip all values in $\mathbf{U}^*$, the value of $X$ will also flip, contradiction. Given the uniformity of $\mathbf{U'}$, the claim follows. We can put this together with the previous claim, and the result follows. We can add fair coins as the input to all other variables outside $F$, which will imply the claim for the whole graph $G$.

In regard to the equality between $I$, note that given that the equality of both models holds for $P$, and removing edges due to interventions will just make some nodes from $\mathbf{U'} \backslash \mathbf{U_w}$ to have an odd number of children, it it not difficult to see based on the previous argument that this just creates more variables that are free to vary, which will entail the same probabilistic uniform behavior in both models. Another way to see this fact is to consider the new exogenous variables from $\{\mathbf{U} \backslash \mathbf{U_w}\}$ that have only one children after the intervention as analogous to $U_w^*$, and so the same argument follows. ∎

Finally, Lemma 2 together with Lemmas 12 and 13 prove Theorem 4. ∎

**Theorem 5** (soundness). *Whenever **sID** returns an expression for $P_x^*(\mathbf{y})$, it is correct.*

*Proof.* Noting that the selection diagram inputted to **sID** is also a causal diagram over $\Pi^*$, and trivial transportability is equivalent to identifiability in $\Pi^*$, the correctness of the identifiability calls was already established elsewhere [27, 34].

It remains to show the correctness of the test in line 10 of **sID**. First note that, by construction, $\mathbf{X'}$ in each local call is always a set of pre-treatment covariates. But now the correctness follows directly by S-admissibility of $\mathbf{X'}$ together with Corollary 1 in Ref. 1. Further note that the set of Z-nodes outside the local component will not affect separability of the S-nodes inside it (following the topology of the hedge), and other S-nodes outside can be removed from the expression before the test. More specifically, note that the effect $Q^*$ in each local call that uses line 10 can be expressed in its expanded form (using a typical C-component decomposition), and given that the independence imposed by S-admissibility holds, together with the fact that both populations share the same causal graph $G$, allow that the functions of $\Pi^*$ to be replaced with the respective functions in $\Pi$, which implies the result. ∎

**Remark 5.** The next results are similar to the identification counterparts given in Refs. 26, 69.

**Theorem 6.** *Assume **sID** fails to transport $P_\mathbf{x}^*(\mathbf{y})$ (executes line 11). Then there exists $\mathbf{X'} \subseteq \mathbf{X}, \mathbf{Y'} \subseteq \mathbf{Y}$, such that the graph pair $D, C_0$ returned by the fail condition of **sID** contain as edge subgraphs sC-forests $F, F'$ that form a s-hedge for $P_{\mathbf{x'}}^*(\mathbf{y'})$.*

*Proof.* Before failure **sID** evaluated false consecutively at lines 5, 6, and 10, so $D$ local to this call is a sC-component, and let $\mathbf{R}$ be its root set. We can remove some directed arrows from $D$ while preserving $\mathbf{R}$ as root, yielding a $\mathbf{R}$-rooted sC-forests $F$. Since by construction $F' = F \cap C_0$ is closed under descendants and only directed arrows were removed, both $F, F'$ are sC-forests. Also by construction, $\mathbf{R} \subset An(\mathbf{Y})_{D_{\overline{\mathbf{x}}}}$ together with the fact that $\mathbf{X}$ and $\mathbf{Y}$ from the recursive call are clearly subsets of the original input, finish the proof. ∎

**Corollary 3** (completeness). ***sID** is complete.*

*Proof.* The result follows from Theorem 6 where $P^*_{\mathbf{x}'}(\mathbf{y}')$ is not transportable in $H$. But now, it is easy to add the remaining variables from $G$, making them independent of $H$ (e.g., as random coins). So, the models in the counterexample induce $G$, and witness the non-transportability of $P^*_{\mathbf{x}}(\mathbf{y})$.

**Corollary 4.** $P^*_{\mathbf{x}}(\mathbf{y})$ *is transportable from* $\Pi$ *to* $\Pi^*$ *in $G$ if and only if there is not s-hedge for* $P^*_{\mathbf{x}'}(\mathbf{y}')$ *in $G$ for any* $\mathbf{X}' \subseteq \mathbf{X}$ *and* $\mathbf{Y}' \subseteq \mathbf{Y}$.

*Proof.* Follows directly from the previous Corollary. ∎

**Theorem 7.** *The rules of do-calculus, together with standard probability manipulations are complete for establishing transportability of all effects of the form* $P^*_{\mathbf{x}}(y)$.

*Proof.* It was shown elsewhere [69] that the steps of **sID** but line 10 correspond to sequences of standard probability manipulations and applications of the rules of *do*-calculus. The line 10 is constituted by a conditional independence judgment, and standard probability operations for the replacement of the functions based on the invariance allowed by the S-admissibility of the local $X'$ in each recursive call (as discussed above in the proof of correctness). ∎